\documentclass{article}

\usepackage{arxiv}

\usepackage[utf8]{inputenc} 
\usepackage[T1]{fontenc}    
\usepackage{hyperref}       
\usepackage{url}            
\usepackage{booktabs}       
\usepackage{amsfonts}       
\usepackage{nicefrac}       
\usepackage{microtype}      
\usepackage{cite}
\usepackage{amsmath,amssymb,amsfonts}
\usepackage{algorithmic}
\usepackage{algorithm}
\usepackage{bm}
\usepackage{amsmath}
\usepackage{graphicx}
\usepackage{textcomp}
\usepackage{xcolor}
\usepackage{todonotes}
\def\BibTeX{{\rm B\kern-.05em{\sc i\kern-.025em b}\kern-.08em
    T\kern-.1667em\lower.7ex\hbox{E}\kern-.125emX}}
\usepackage{multicol}
\newcommand{\K}{\bm{K}}

\title{Scalable Hyperparameter Optimization with Lazy Gaussian Processes}

\author{Raju Ram$^{1,3}$\thanks{Raju Ram and Sabine M\"uller equally contribute to this paper.}, Sabine M\"uller$^{1,2}$, Franz-Josef Pfreundt$^{1,2}$, Nicolas R. Gauger$^3$, Janis Keuper$^{1,4}$\\[5mm]
$^1$Competence Center High Performance Computing, Fraunhofer ITWM, Kaiserslautern, Germany\\
$^2$Fraunhofer Center Machine Learning, Germany\\
$^3$Scientific Computing Group, TU Kaiserslautern, Germany\\
$^4$Institute for Machine Learning and Analytics, Offenburg University, Germany\\
\{raju.ram, sabine.b.mueller\}@itwm.fraunhofer.de\\
}

\begin{document}
\maketitle

\begin{abstract}
Most machine learning methods require careful selection of hyper-parameters in order to train a high performing model with good generalization abilities. Hence, several automatic selection algorithms have been introduced to overcome tedious manual (try and error) tuning of these parameters. Due to its very high sample efficiency, Bayesian Optimization over a Gaussian Processes modeling of the parameter space has become the method of choice. Unfortunately, this approach suffers from a cubic compute complexity due to underlying Cholesky factorization, which makes it very hard to be scaled beyond a small number of sampling steps. \\
In this paper, we present a novel, highly accurate approximation of the underlying Gaussian Process. Reducing its computational complexity from cubic to quadratic allows an efficient strong scaling of Bayesian Optimization while outperforming the previous approach regarding optimization accuracy. The first experiments show speedups of a factor of 162 in single node and further speed up by a factor of 5 in a parallel environment.
\end{abstract}


\section{Introduction}
\begin{figure}[t!]
  \centering
  \includegraphics[scale=0.55]{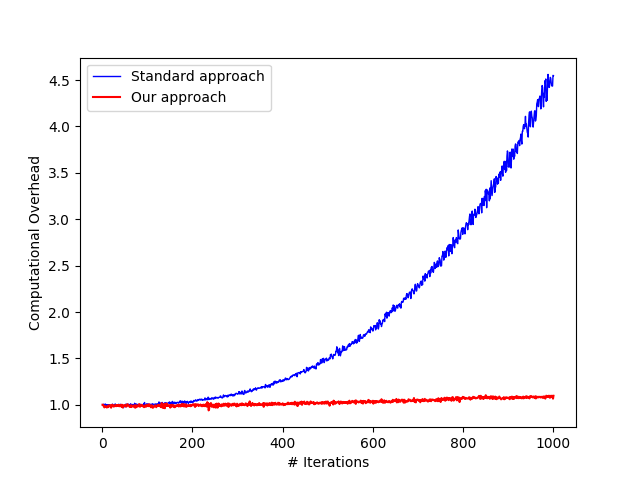}
  \caption{Comparison of the computational overheads of the original Bayesian hyper-parameter optimization and our approach. Training of LeNet with MNIST data optimizing 5 parameters.}
  \label{fig:mnist_comp_overhead}
\end{figure}
Deep neural networks constitute the state of the art in many research areas. Especially the field of image analysis is dominated by deep learning approaches ranging from image classification, over-segmentation, to tracking.\\
In addition to model design, one of the main challenges is the selection of hyperparameters. These parameters have a heavy impact on nearly all parts of the deep learning model: From training parameters such as learning rate, weight decay or momentum, up to model decisions itself a good parameter choice is essential.\\
The optimal parameter selection is of crucial importance - while being extremely difficult and time-consuming. Endless trial and error procedures are not the exception but rather the common practice.
One trivial strategy to find well-suited parameters is grid search, i.e. exploration of all possible parameter combinations. Although simple, its results are usually not satisfying \cite{bergstra2012random}.  It turned out, that random search is a more efficient way to elaborate parameter settings, considering only a subset of possible configurations \cite{bergstra2012random}. \\
However, these approaches are still not sample efficient and hyperparameter optimizations do neither trace a reasonable strategy nor do they remember past evaluations.\\
One possible approach to automate parameter selection is Bayesian optimization \cite{snoek2012practical,eggensperger2013towards, feurer2015initializing}. 
Such algorithms follow the strategy to create a probability model of the objective function and select the most promising hyperparameters to be evaluated in the target function. Among these optimization techniques, the most common approach is based on Gaussian processes \cite{williams1996gaussian, rasmussen2006gaussian}. \\
Gaussian processes represent a very powerful approach to hyperparameter optimization - but they are also very popular in other areas, such as regression and extrapolation \cite{gardner2018product}. Typically, the most important limitation is its computational complexity. The most computationally intensive step is the inversion of a quadratic matrix, which is necessary for the solution of a linear system of equations as well as log determinants. Usually, this inversion is carried out with the help of a Cholesky factorization. Nevertheless, the computational effort increases cubically in the number of matrix entries.\\
Neural architecture search is currently one of the most active research fields in the area of deep learning \cite{real2019regularized,elsken2018neural,zoph2018learning}. Here, complete architectures are automatically optimized taking the human out of the loop. The network search operating in a nearly infinitely dimensional space has a tremendous demand concerning computational time, as well as computing power. 
In addition to the general hyper-parameters which have to be optimized individually for each neural network, the individual network layers, as well as their connections, are also optimized. In this scenario, the cubic complexity of the Gaussian process inference is an even more serious problem: To find well-performing architectures, often tens of thousands of iterations of topology updates are necessary - massively hitting the bottleneck of the matrix inversions \cite{klein2018towards}.
\section{Related Work}
Nowadays there are various inference techniques for Gaussian processes. Recently so-called MVM methods (Matrix-Vector-Multiplications) have become very popular. These iterative approaches use Krylov subspace methods and reduce operations related to the covariance matrix to matrix-vector multiplications. Some of these approaches have low runtimes but introduce a new problem: The input dimension is strongly limited \cite{wilson2015kernel,gardner2018product}. Especially in the setting of neural network hyperparameter optimization, the input dimension cannot be limited - the number of parameters to be optimized strongly depends on the considered architecture and can increase dramatically. Hence, these approaches are not applicable.\\
Alternatively, it is possible to divide the data into subsets to distribute computation among independent computational units. These local experts then calculate an approximation of the distribution, while all of them have their collection of hyperparameters that need to be optimized \cite{deisenroth2015distributed,nguyen2014fast}.\\
Another class of approaches is based on inducing point methods. Here a rank $m$ approximation of the covariance matrix is calculated with the help of a subset $m < n$ \cite{gardner2018product,hensman2013gaussian}. However, their computational efficiency is limited. 

Our contributions are as follows: We address the limiting complexity of Bayesian Optimization from a high-performance perspective: In a first step, we relax the matrix inversion in an intuitive and highly efficient way. Afterward, we show that with an adaptive lagged update of the matrix inversion, we solve different computational challenges in less time while not losing accuracy. 
\subsection{Organization of the Paper}
In Section \ref{sec:method}, we introduce the underlying algorithm of Bayesian optimization, explain its structure and propose our novel method\footnote{code available at: \url{https://github.com/cc-hpc-itwm/HPO_LazyGPR}} to Bayesian optimization with lazy Gaussian processes. Section \ref{sec:experiments} presents experimental analyses of both approaches in 4 different settings. We conclude our work with a summary and outlook in Sec. \ref{sec:conclusions}.

\section{Method}
\label{sec:method}
\begin{figure*}[ht!]
  \centering
  \includegraphics[scale=0.49]{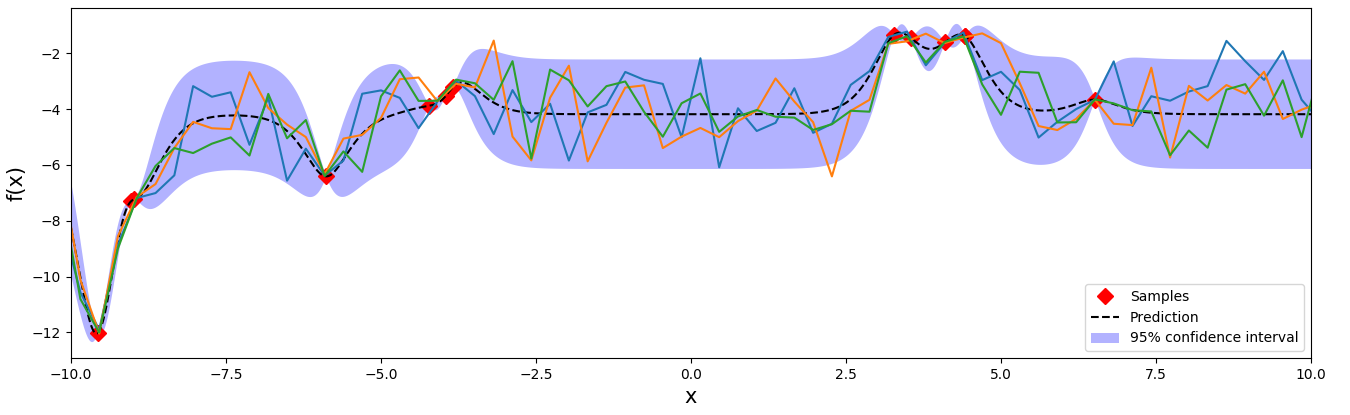}
  \caption{1D Levy function with 12 random seed points.}
  \label{fig:intro_bayes}
\end{figure*}

To describe our approach, we first discuss the general concepts of Bayesian Optimization \cite{kushner1964new} and Gaussian processes. Afterward, we address the computationally most expensive part, the Cholesky decomposition. Finally, we explain our modifications to lazy Gaussian processes.
\subsection{Bayesian Optimization}
A Bayesian optimization is a powerful approach for the optimization of multimodal target functions that are expensive to evaluate. One of its core aspects is sample efficiency - with just a few evaluations it is usually possible to find the optimal solution. It is therefore desirable to be able to use the method for the determination of hyperparameters in neural networks.\\
Let us consider some training data $\mathcal{D}_{train}$ and test data $\mathcal{D}_{test}$, as well as their validation error $\mathcal{V}(\bm{x}, \mathcal{D}_{train}, \mathcal{D}_{test})$ with respect to a specific parameter setting $\bm{x}$.
We follow Feurer et al. \cite{feurer2015initializing} and define the hyperparameter optimization problem as
\begin{align}
f^\mathcal{D}(\bm{x}) = \frac{1}{k}
\sum_{i=1}^k \mathcal{V}(\bm{x}, \mathcal{D}^{(i)}_{train}, \mathcal{D}^{(i)}_{test}).
\end{align}
Here, $x_1, \dots x_n$ are the hyperparameters to be optimized and $\Theta_1 \dots \Theta_n$ are their domains, respectively. The corresponding search space is then denoted as $\Theta_1 \times \dots \Theta_n$ and we use $k$-fold cross validation.\\
The goal is now to find an approximated optimum for the computational expensive training of a neural network. Similar to a normal function, the network accepts an input vector $\bm{x} \in \mathbb{R}^d$ of parameters and returns a scalar number - in our case the accuracy. \\
The cornerstone of Bayesian optimization is the interplay of the surrogate model, incorporating beliefs about the target function, and the acquisition function, proposing where to draw the next sample. For simplicity, let us consider the one-dimensional case for an unknown function $f$, see Fig. \ref{fig:intro_bayes}. At the beginning we only know the function evaluations $f(\bm{x})$. We can then apply Bayes rule to get the posterior Gaussian process, i.e. our surrogate model.

\subsection{Gaussian Processes}
A Gaussian process (GP) is a probability distribution over functions \cite{williams1996gaussian}. Here, it is restricted to the space of functions that fit the already evaluated sample positions. The predicted function corresponds to the mean, while the standard deviation describes the possible deviation of functions from its mean.\\
In other words, a GP is a random process in which each point $x \in \mathbb{R}^d$ is mapped to a random variable $f(x)$. The joint distribution of these variables $p(f(x_1), \dots, f(x_i))$ - 
where $x_i, i=1, \dots n$ is finite - is itself a Gaussian:
\begin{align}\label{eq:prior_gp}
    p(f | \mathcal{X}) = \mathcal{N}(f~|~\mathbf{\mu}, \mathbf{K})).
\end{align}
Here, $\mu=(\mu(x_1), \dots, \mu(x_n))$ is the mean function, and K the symmetric semi-definite covariance matrix, defined as $\K_{i,j} = \kappa (x_i, x_j)$. The positive semi-definite kernel function $\kappa$ guarantees a minimal distance in output space when two input values are similar. A common choice here are Matern kernels for twice differentiable functions \cite{rasmussen2006gaussian}:
\begin{align}
    \kappa_{2.5}(d)=\sigma ^{2}
    {\Bigg (}{1 + \frac{\sqrt{5}d}{\rho} + \frac{5d^2}{3\rho^2}}
    {\Bigg )} \exp {\Bigg (}{\frac{\sqrt{5}d}{\rho}}{\Bigg )}
\end{align}
where $d$ is the distance between two points, and $\rho$ is the length scale. Typically, Bayesian optimization strategies exploit the iterative manner of updating the Gaussian posterior to learn the kernel parameters. We refrain from this strategy to enable reusing previously computed covariance matrices and fix $\rho=1$, see Sec. \ref{sec:lgp}.
After gathering some sample evaluations $y=f(x)$, it is possible to turn the prior GP of \eqref{eq:prior_gp} into its posterior counterpart \cite{jordan2015machine}
\begin{align}
\begin{split}
    p(f_*~|~\mathcal{X}_*, \mathcal{X}, \mathbf{y}) = \int  p(f_*~|~\mathcal{X}_*,f)p(f~|~\mathcal{X},\mathbf{y}).
\end{split}
\end{align}
Following the definition of GP, we can write the joint distribution of observing $\mathbf{f}_*$ when $\mathbf{y}$ is given as \cite{jordan2015machine}
\begin{align}
    \begin{pmatrix}
    \mathbf{y} \\
    \mathbf{f}_*
    \end{pmatrix}
    =
    \mathcal{N}
    \begin{pmatrix}
    \begin{pmatrix}
    \mathbf{\mu} \\
    \mathbf{\mu}_*
    \end{pmatrix},
    \begin{pmatrix}
    \mathbf{K}_y & \mathbf{K}_* \\
    \mathbf{K}^T_* & \mathbf{K}_{**}
    \end{pmatrix}
    \end{pmatrix}
\end{align}
where $\mathbf{K}_y = \kappa(\bm{x}, \bm{x}) + \sigma^2\mathbf{I}$, $\mathbf{K}_* = \kappa(\bm{x}, \bm{x}_*)$, and $\mathbf{K}_{**} = \kappa(\bm{x}_*, \bm{x}_*)$ with $\mathbf{K}_y \in \mathbb{R}^{N\times N}$, $\mathbf{K}_* \in \mathbb{R}^{N\times N_*}$, and $\mathbf{K}_{**} \in \mathbb{R}^{N_*\times N_*}$.
Using conditioning, we can write the posterior now as
\begin{align}
\begin{split}
     p(f_*~|~\mathcal{X}_*, \mathcal{X}, \mathbf{y})&= \mathcal{N}(f_*~|~\mathbf{\mu}_*,\mathbf{\Sigma}_*))\\
    \text{where} \  \mu_* &= \mathbf{K}^T_*\mathbf{K}^{-1}_y \mathbf{y}, \\
    \Sigma_* &= \mathbf{K}_{**} -   \mathbf{K}^T_* \mathbf{K}^{-1}_y \mathbf{K}_*.
\end{split}
\end{align}
We can then summarize the procedure to estimate the prediction as well as the log marginal likelihood as Alg. \ref{algo:gpr} \cite{williams1996gaussian}.
\begin{algorithm}[b!]
 \caption{Bayesian Optimization: General algorithm for prediction and log marginal likelihood estimation.}
 \label{algo:gpr}
\begin{algorithmic}[1]
  \STATE \emph{Input}: $\mathcal{X},  \bm{y}, k, \sigma^2, \bm{x_*}$
  \smallskip
  \STATE $\mathbf{L} = \text{cholesky}(\mathbf{K} + \sigma^2\mathbf{I})$
  \smallskip
  \STATE $\bm{\alpha} =  \mathbf{L}^{-T}(\mathbf{L}^{-1} \mathbf{y})$
  \smallskip
  \STATE $\bar{f}_* = \bm{k_*}^T \bm{\alpha}$
  \smallskip
  \STATE $\bm{v} = \mathbf{L}^{-1}\bm{k_*}$
  \smallskip
  \STATE $\mathbb{V}[f_*] = k(\bm{x_*}, \bm{x_*}) - \bm{v}^T\bm{v}$
  \smallskip
  \STATE $\log p(\bm{y}|\mathcal{X}) = -\frac12 \bm{y}^T\bm{\alpha} - \sum_i \log \mathbf{L}_{ii} - \frac{n}{2} \log 2\pi$
 \end{algorithmic}
 
\end{algorithm}
Here, $\bar{f}_*$ is the mean, $\mathbb{V}[f_*]$ the variance and $\log p(y|\mathcal{X})$ the log marginal likelihood. Unfortunately, the computational complexity of the Cholesky factorization in line 2 is $\mathcal{O}(n^3)$, limiting its application.\\
Fig. \ref{fig:intro_bayes} shows an visual example: In total we draw 12 samples from a 1D Levy function, defined
as 
\begin{align}
\begin{split}
 f_L(x) = &\sin^2(\pi w) + (w-1)^2  [1+sin^2 (2\pi w)]  \\ 
    &\text{where} \ w = 1 + \frac{x- 1 }{4}.
\end{split}
\end{align}
Let us assume in the following, that we want to find
\begin{align}
    \max_x(- f_L(x)).
\end{align}
The blue shaded area denotes the standard deviation or the space of possible functions. We draw 3 functions from the multivariate distributions, shown in dark blue, orange, and green. All of them fit the present function evaluations and lie in this space. The dashed line marks the mean value of the GP posterior. Adding a new function evaluation allows refining the GP posterior. Please note that the standard deviation is higher, the larger the distance to already seen samples is: The Gaussian process can identify regions of uncertainty. \\
Looking at the GP posterior, the question arises where to evaluate next. There are two main aspects to be considered:
\begin{enumerate}
\item We should evaluate points at which high values are predicted (exploration),
\item we should especially check areas of uncertainty and reduce the variance in the assumption (exploitation).
\end{enumerate}
We can represent our preference in this trade-off with an acquisition (or utility) function. This function, typically cheap to evaluate, allows us to predict which position is particularly suitable for the next evaluation. The most frequently used acquisition function is expected improvement \cite{movckus1975bayesian} - However, exchanging the utility function does not influence the overall structure. We focus in the following on expected improvement, but without loss of generality.

\subsubsection{Expected Improvement}\label{sec:ei}
\begin{figure*}[h!]
  \centering
  \includegraphics[scale=0.49]{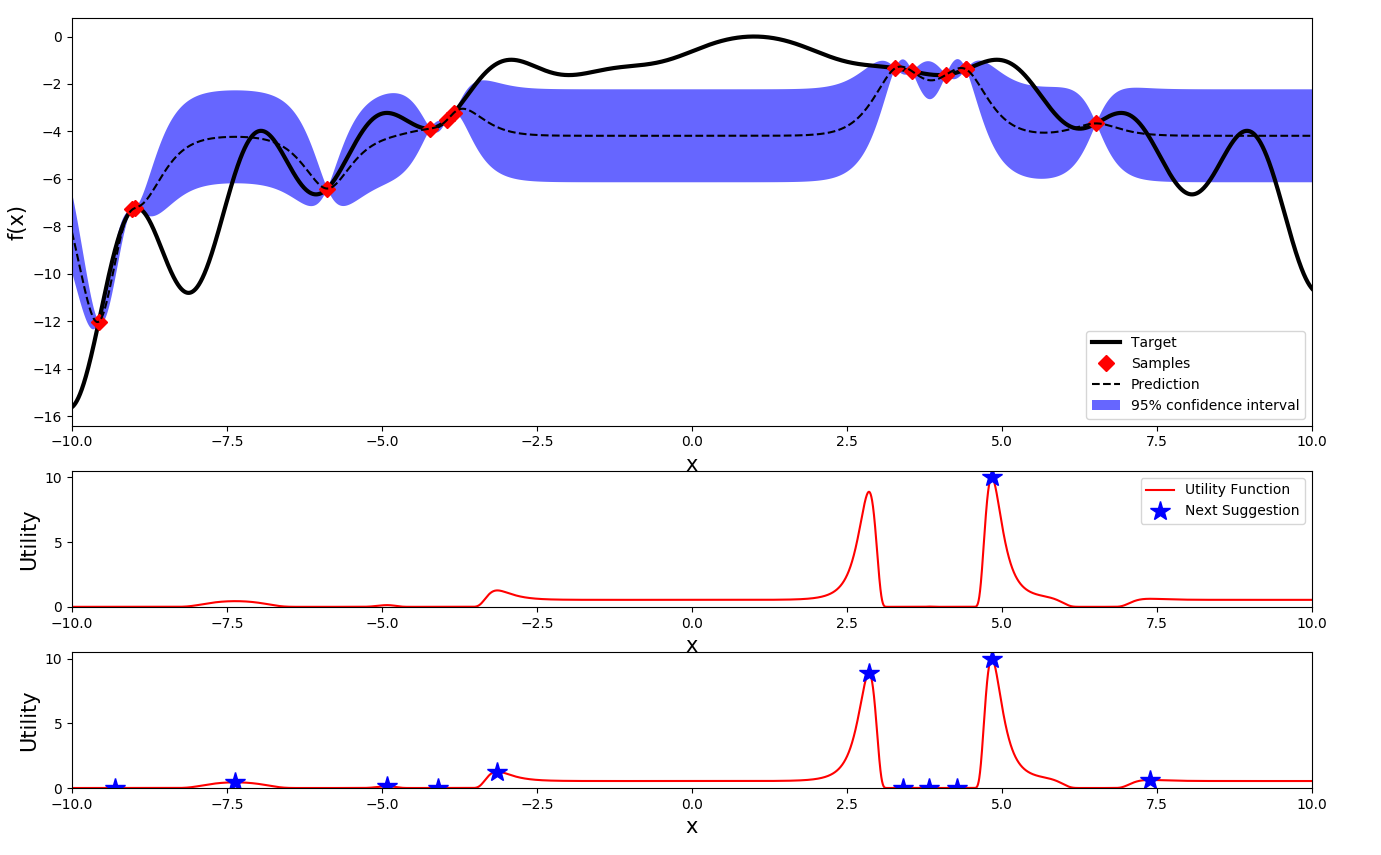}
  \caption{1D Levy function with 12 random seed points. (top) Gaussian process, (middle) suggestion of standard expected improvement, (bottom) suggestions for all local maxima of expected improvement. Utility functions are re-scaled for visibility.}
  \label{fig:parallel_suggestions}
\end{figure*}

The acquisition function of expected improvement suggests the parameter set $\bm{x}_{n+1}$ where the largest improvement is to be expected \cite{movckus1975bayesian,frazier2018tutorial}. \\
Let $n$ function evaluations of the surrogate model be given and $f'_n = \max_{m\leq n} f(\bm{x}_m)$ be the maximal value we found so far.
Then the expected improvement of evaluation for any new point $\bm{x}_{n+1}$ is 
\begin{align}
    \text{EI}_{n+1}= \mathbb{E}[\max(f(\bm{x}_{n+1})-f'_n, 0)].
\end{align}
The suggested point to evaluate next is then given as
\begin{align}
    \bm{x}_{n+1} = \text{argmax} \text{ EI}_{n+1}(\bm{x}).
\end{align}
The expected improvement can be evaluated under the GP model as \cite{movckus1975bayesian,jones1998efficient}
\begin{align}
\begin{split}
    \text{EI}_{n+1} &= 
\begin{cases}
     \gamma\Phi(Z)+ \sigma(\bm{x})\phi(Z),& \text{if } x\geq \sigma(\bm{x}) >0\\
    0,              & \text{otherwise}
\end{cases}\\
\text{where}\\ 
\gamma &= (\mu(\bm{x}) - f(\bm{x}_{n+1}) - \xi),\\
Z &= 
\begin{cases}
    \frac{\gamma}{\sigma(\bm{x})},& \text{if } \sigma(\bm{x}) >0\\
    0,              & \text{otherwise}
\end{cases}
\end{split}
\end{align}
Here, $\mu$ and $\sigma$ are the mean and standard deviation of the GP posterior. The trade-off parameter $\xi$ defines the amount of exploration, i.e. the influence of the posterior mean $\mu$ decreases as $\xi$ increases. Fortunately, EI is continuous differentiable and standard techniques can be applied. Typically, the optimal solution is found via initialization with different seed points and several restarts of the optimization process. The maximal value is then suggested as next sample position, see Fig. \ref{fig:parallel_suggestions}(middle).

\subsection{Lazy Gaussian Processes}\label{sec:lgp}
The main bottleneck in Gaussian process inference is the computation of the Cholesky decomposition for the covariance matrix $\K$. We denote covariance matrix for $n$ existing sample points by $\K_n$. The Cholesky factorization is the unique decomposition of a symmetric and positive-definite matrix $\K_n \in \mathbb{R}^{n \times n} $ into the product \cite{williams1996gaussian,gill1974newton}
\begin{align}
    \K_n =  \mathbf{L}_n\mathbf{L}_n^T.
\end{align}
where $\mathbf{L}_n$ is a lower triangular matrix with real and positive diagonal entries. Unfortunately, the full Cholesky decomposition (see Alg. \ref{algo:Cholesky}) has an asymptotically complexity of $\mathcal{O}(\frac{n^3}{6})$, limiting the application of Gaussian processes. We have removed the subscript $n$ from $\K_n$ in Alg. \ref{algo:Cholesky} for better notation. \\
In the standard approach of Bayesian optimization with Gaussian processes, the parameters of the covariance kernel are permanently updated, leading to changing covariance matrices $\K$ as soon as new samples are drawn. \\
\begin{algorithm}[t]
\caption{Cholesky factorization of a matrix $\K \in \mathbb{R}^{n \times n}$}  
\label{algo:Cholesky}
\begin{algorithmic}[1]  
\FOR{ $i=1$ to $n$}
        \FOR{ $j=1$ to $i-1$}
            \FOR{ $k=1$ to $j-1$}
            \STATE $\K_{i,j} =  \K_{i,j} - \K_{i,k} * \K_{j,k}$
            \ENDFOR
            \STATE $\K_{j,j} =  \K_{i,j} / \K_{j,j}$
            \ENDFOR
            \FOR{ $\K=1$ to $i-1$}
            \STATE $\K_{i,i} =  \K_{i,i} - \K_{i,k}^2$
            \ENDFOR
        \STATE $\K_{i,i} = \sqrt{\K_{i,i}}$
\ENDFOR

\FOR{ $i=1$ to $n$}
    \FOR{ $j=i+1$ to $n$}
      \STATE $\K_{i,j} = 0$  \qquad \COMMENT{Set upper triangular part of $\K$ to $0$}
    \ENDFOR
\ENDFOR
\STATE \textbf{return} $\K$
\end{algorithmic}  

\label{chol_algo}  
\end{algorithm}
However, if we make an approximation in which we do not update the kernel parameters, the new co-variance matrix  $\K_{n+1} \in \mathbb{R}^{(n+1) \times (n+1)}$ has the same entries in its $n \times n$ sub-matrix. In particular, only a new column vector $\mathbf{p} \in \mathbb{R}^n $, a new row vector $\mathbf{p}^T \in \mathbb{R}^n $ as well as a value $c \in \mathbb{R}$ are added resulting in the following matrix structure
\begin{align}
\label{eqn:3B_2}
\K_{n+1} =
    \begin{pmatrix}
    \mathbf{K}_n & \mathbf{p} \\
    \mathbf{p}^T & c
    \end{pmatrix}
\end{align}
Obviously $\K_{n+1}$ is only an extension of $\K_n$, such that we do not have to factorize $\K_{n+1}$ completely. Instead, we can reuse the lower triangular matrix $ \mathbf{L}_n$ from the previous factorization, where $\K_n = \mathbf{L}_n\mathbf{L}_n^T$, so that $\K_{n+1} = \mathbf{L}_{n+1}\mathbf{L}_{n+1}^T$, where
 \begin{align}
 \label{eqn:3B_3}
 \mathbf{L}_{n+1} = 
 \begin{pmatrix}
 \mathbf{L}_n & \mathbf{0} \\
 \mathbf{q}^T & d
 \end{pmatrix} \qquad
 \mathbf{q} \in \mathbb{R}^n, 
 d \in \mathbb{R}
 \end{align}
According to the definition of the Cholesky factorization, the following holds:
 \begin{align}
 \label{eqn:3B_3}
 \mathbf{L}_{n+1} = 
 \begin{pmatrix}
 \mathbf{L}_n & \mathbf{0} \\
 \mathbf{q}^T & d
 \end{pmatrix} \qquad
 \mathbf{q} \in \mathbb{R}^n, 
 d \in \mathbb{R}
 \end{align}
According to the definition of the cholesky factorization, the following holds:
\begin{align}
\label{eqn:3B_4}
\begin{split}
\begin{pmatrix}
\mathbf{K}_n & \mathbf{p} \\
\mathbf{p}^T & c
\end{pmatrix}
&=
\begin{pmatrix}
\mathbf{L}_n & \mathbf{0} \\
\mathbf{q}^T & d
\end{pmatrix}
\begin{pmatrix}
\mathbf{L}_n^T & \mathbf{q} \\
\mathbf{0} & d
\end{pmatrix} \\[1mm] 
&=
\begin{pmatrix}
\mathbf{L}_n\mathbf{L}_n^T & \mathbf{L}_n\mathbf{q} \\
\mathbf{q}^T\mathbf{L}_n^T & \mathbf{q}^T\mathbf{q}+d^2
\end{pmatrix}
\end{split}
\end{align}
Comparing the left and right hand side from \eqref{eqn:3B_4}, we obtain \eqref{eqn:3B_5}. Here, we compute $\mathbf{q}$ and  $d$ in order to get a new row in $\mathbf{L}_{n+1}$ as follows:
\begin{equation}
\label{eqn:3B_5}
    \begin{aligned}
  \mathbf{L}_n\mathbf{q} = \mathbf{p}  \\ 
    \quad d=\sqrt{c- \mathbf{q}^T\mathbf{q}} 
    \end{aligned}
\end{equation}
We obtain $\mathbf{q}$ by solving the system of equations $\mathbf{L}_n\mathbf{q} = \mathbf{p}$ using forward substitution in a time complexity of $\mathcal{O}(n^2)$. Computation of $d$ involves a dot product of $\mathbf{q}$ with a time complexity of $\mathcal{O}(n)$. Therefore $ \mathbf{L}_{n+1}$ is obtained in an asymptotic time of $\mathcal{O}(n^2) + \mathcal{O}(n) = \mathcal{O}(n^2)$ instead of the  $\mathcal{O}(n^3)$ complexity for na\"ive Cholesky factorization.\\
In the similar way, the vector $\mathbf{q}$ and the diagonal entry $d$ are calculated for successive samples (i.e. for iteration $i$) using the lower triangular matrix, computed previously. Our approach is as follows: in the first iteration, we calculate a complete Cholesky decomposition to obtain $\mathbf{L}_{n+1}$. For all further iterations, we calculate only the new vector $\mathbf{q}$ and the diagonal entry $d$. These are then used to form the new Cholesky factor $\mathbf{L}_{n+i}$ using the previous Cholesky factor $\mathbf{L}_{n+i-1}$ for a new sample number $i$; 
see Alg. \ref{Choleskyoptimized_algo}. 
\begin{algorithm}[t]
\caption{Iterative Cholesky factorization for $N$ samples points (iterations) with $\bm{x_1}, \bm{x_2}, \ldots , \bm{x_n}$ seeds}  
\label{Choleskyoptimized_algo}
\begin{algorithmic}[1]  
\IF {$N=1$}
    \STATE $\mathbf{p}^T := [  k(\bm{x_{1*}}, \bm{x_1}) \ k(\bm{x_{1*}}, \bm{x_2}) \  \ldots  \ k(\bm{x_{1*}}, \bm{x_n}) ] $  \smallskip
    \STATE c := $  k(\bm{x_{1*}}, \bm{x_{1*}})$  \smallskip
    \STATE
             $\K_{n+1} = 
             \begin{pmatrix}
              \mathbf{K}_n & \mathbf{p} \\
              \mathbf{p}^T & c
              \end{pmatrix}$
             \smallskip
    \STATE Compute Cholesky factor $\mathbf{L}_{n+1}$ using Alg. \ref{algo:Cholesky} with matrix $\K_{n+1}$ 
\ELSE
    \FOR{ $i=2, 3, \ldots , N$}
        \STATE $\mathbf{p}^T := [  k( \bm{x_{i*}}, \bm{x_1}) \ k(\bm{x_{i*}}, \bm{x_2}) \  \ldots \ k(\bm{x_{i*}}, \bm{x_{n+i-1}} )  ] $  \smallskip
        \STATE $c :=  k(\bm{x_{i*}}, \bm{x_{i*}})$   \smallskip
        \STATE 
                $\K_{n+i} = 
                \begin{pmatrix}
                 \mathbf{K}_{n+i-1} & \mathbf{p} \\
                 \mathbf{p}^T & c
                 \end{pmatrix}$\smallskip
        \STATE Solve for $\mathbf{q}$ in $\mathbf{L}_{n+i-1} \mathbf{q} = \mathbf{p}$ \smallskip
        \STATE $d=\sqrt{c- \mathbf{q}^T\mathbf{q}}$  \smallskip
        \STATE 
                $\mathbf{L}_{n+i} =
                \begin{pmatrix}
                 \mathbf{L}_{n+i-1} & \mathbf{0} \\
                 \mathbf{q}^T & d
                 \end{pmatrix}$
    \ENDFOR
\ENDIF
\RETURN  $\mathbf{L}_{n+N}$
\end{algorithmic}  
\label{chol_algo_opt}  
\end{algorithm}\\
\textbf{Lemma: $d$ is well defined.} For $d$ to be well defined, the term inside the square root, i.e. $c- \mathbf{q}^T\mathbf{q}$ has to be positive. Using \eqref{eqn:3B_5}, we write
\begin{align*}
    \begin{aligned}
  \mathbf{q}=\mathbf{L}_n^{-1}\mathbf{p} \ \text{from} \  \mathbf{L}_n\mathbf{q}=\mathbf{p} \\ 
    \quad d=\sqrt{c- \mathbf{q}^T\mathbf{q}}  \\ 
      d =  \sqrt{c- \mathbf{p}^T (\mathbf{L}_n^{-1})^T \mathbf{L}_n^{-1} \mathbf{p}} \\
         d = \sqrt{c- \mathbf{p}^T \K_n^{-1} \mathbf{p}}
    \end{aligned}
\end{align*}
We further reformulate exploiting symmetric positive definiteness of $\K_{n+1}$ where 
\begin{align*}
         \begin{pmatrix}
          \mathbf{I} & \mathbf{0} \\
          -\mathbf{p}^T\mathbf{K}_n^{-1} & 1
          \end{pmatrix}
    \end{align*}

\begin{align}
        Y := X\K_{n+1}X^T = 
        \begin{pmatrix}
\mathbf{K}_{n} & \mathbf{0} \\
\mathbf{0} & c-\mathbf{p}^T\mathbf{K}_n^{-1}\mathbf{p}
\end{pmatrix}
\end{align}
With Sylvester's Inertia Theorem \cite{SylvesterTheorm}, the matrix $Y$ has the same inertia as the matrix $\K_{n+1}$. Consequently, the matrix $Y$ has $n$ positive eigenvalues (since $\K_{n}$ is symmetric positive definite) and one eigenvalue with the same sign as that of $c - \mathbf{p}^T  \K_{n}^{-1} \mathbf{p}$. Since all the eigenvalues in $\K_{n+1}$ are positive, the term $c - \mathbf{p}^T  \K_n^{-1} \mathbf{p}$ is also positive. Therefore, $ d = \sqrt{c- \mathbf{p}^T \K_{n}^{-1} \mathbf{p}}$ is well defined.

\subsection{Parallelization}\label{sec:parallelization}
In the standard approach for Bayesian optimization, the matrix inversion via Cholesky factorization is a tremendous bottleneck. However, in our approach, this problem is nearly resolved such that we can reduce the computational overhead dramatically by iterative computation of the Cholesky factors. It is now straight forward to not only evaluate the best suggestion of the acquisition function but to assess the function values at all local maxima, see Fig. \ref{fig:parallel_suggestions} (bottom). \\
This gives us a dramatic advantage especially in the case of neural network hyperparameter optimization and neural architecture search. We can train $t$ neural network architectures in parallel and synchronize their results easily via iterated computation of the new Cholesky factors resulting in computational costs of $t\mathcal{O}(n^2)$ per iteration.

\section{Experiments}
\label{sec:experiments}

We evaluate our approach in three different settings. First, we address the optimization of the $d$-dimensional Levy-function for 1000 iterations. Second, we optimize the hyperparameters of LeNet for classification of the MNIST data set again for 1000 iterations. Although very simple, this classic challenge allows investigating the behavior of our optimization strategy. Last but not least, we optimize the hyperparameters in the more realistic scenario of Resnet32 for CIFAR10 classification - first sequentially, then in a parallelized approach, each for 300 iterations.\\
For all our experiments, we use 3-fold cross-validation and conducted them on a GPU cluster with 16 compute nodes each with AMD Threadripper 1920x with 3.5GHz and 12 cores, and 64GB RAM. At each node, two NVIDIA GeForce GTX1080Ti with 11GB GDDR5X RAM and 3584 CUDA cores with 1480MHz are installed. The training of the neural networks was performed with TensorFlow 1.12 \cite{tensorflow2015-whitepaper}. In the case of our parallel experiment, we used in total 20 GPUs on 10 nodes.
\subsection{Levy Function}
\begin{figure}[b!]
  \centering
  \includegraphics[scale=0.18]{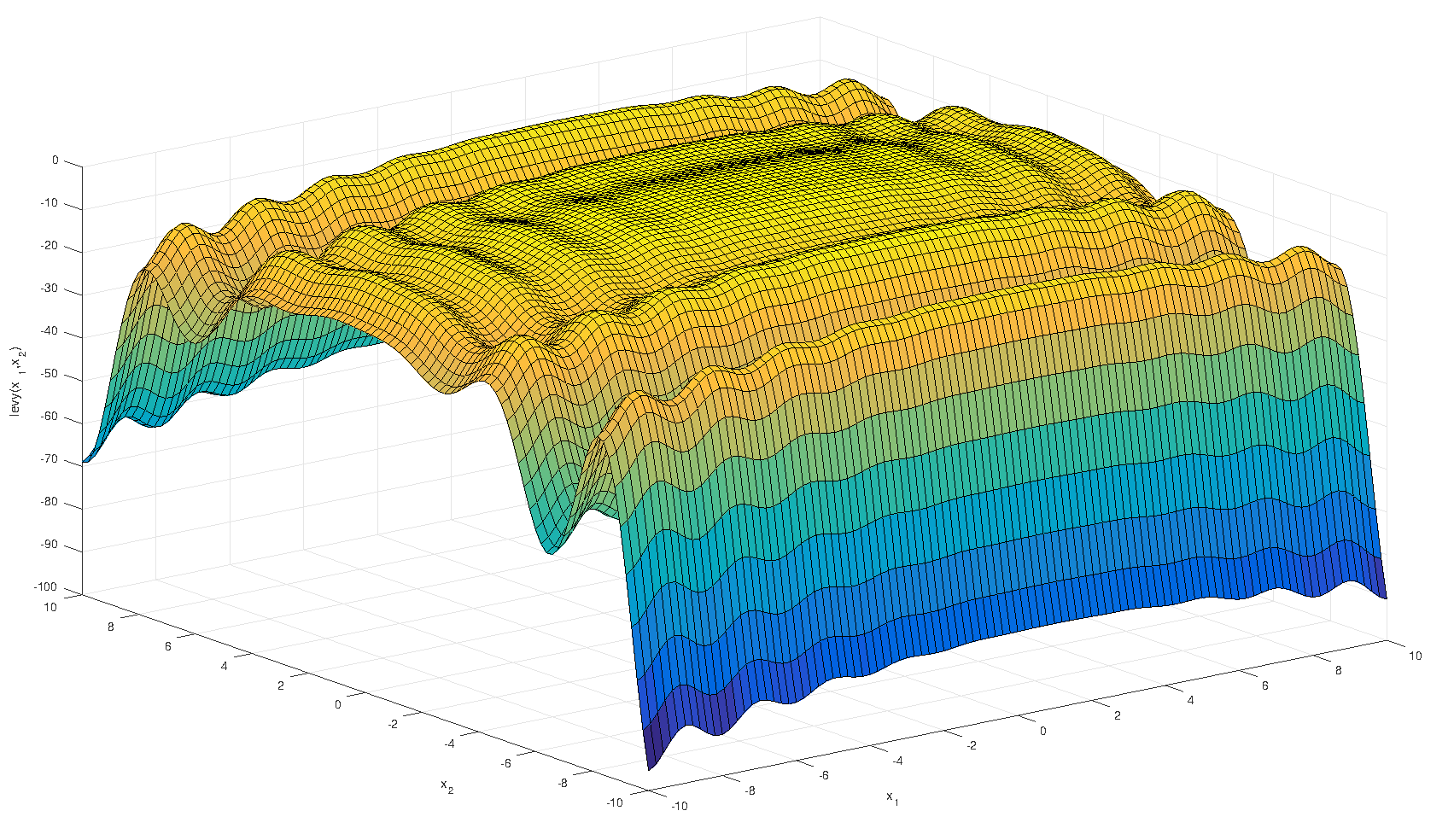}
  \caption{$2$ dimensional negative levy function illustrating the complexity of the function.}
  \label{fig:levy}
\end{figure}

\begin{figure}[t!]
  \centering
  \includegraphics[scale=0.7]{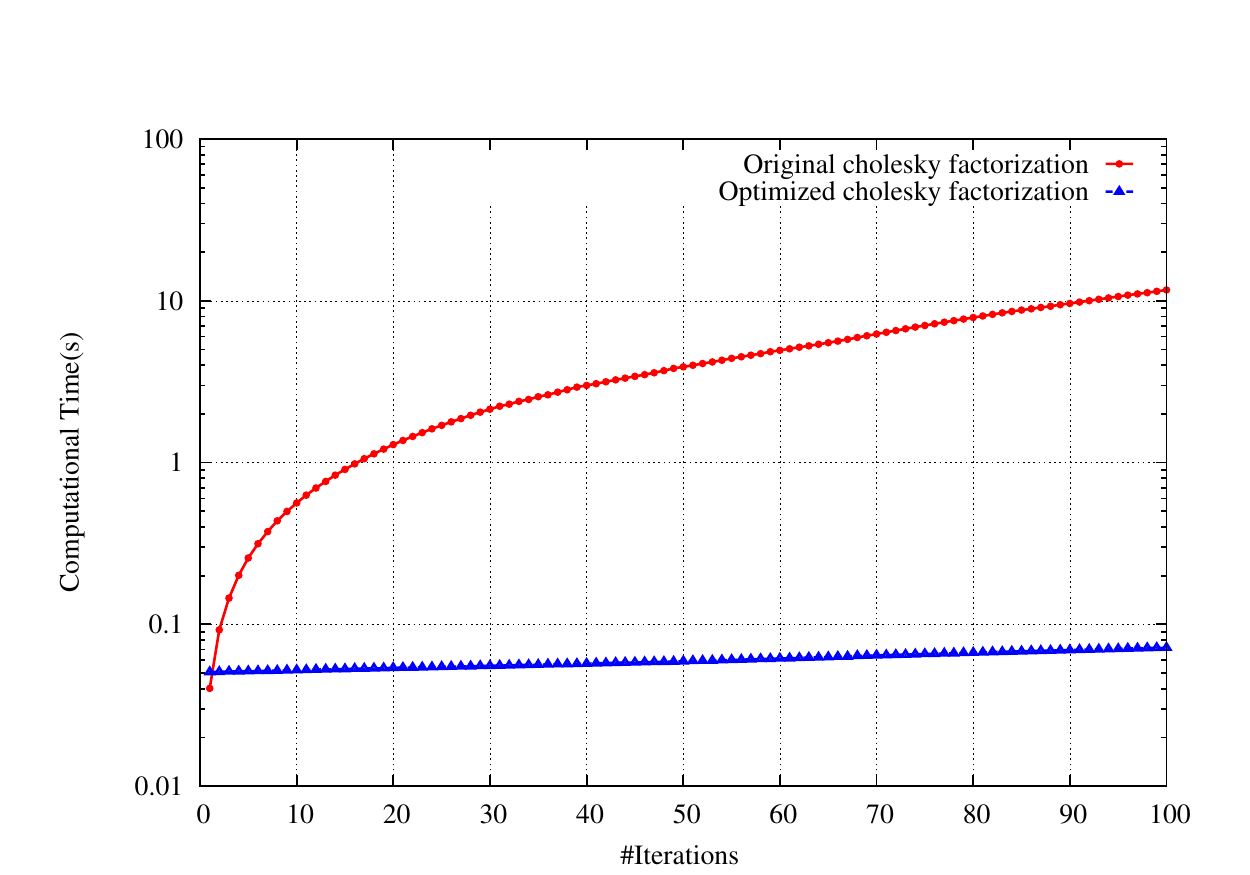}
  \caption{Comparison of computational time of original Cholesky factorization and our approach}
  \label{fig:chol_vs_iter}
\end{figure}

\begin{figure*}[t!]
  \centering
  \includegraphics[scale=1.3]{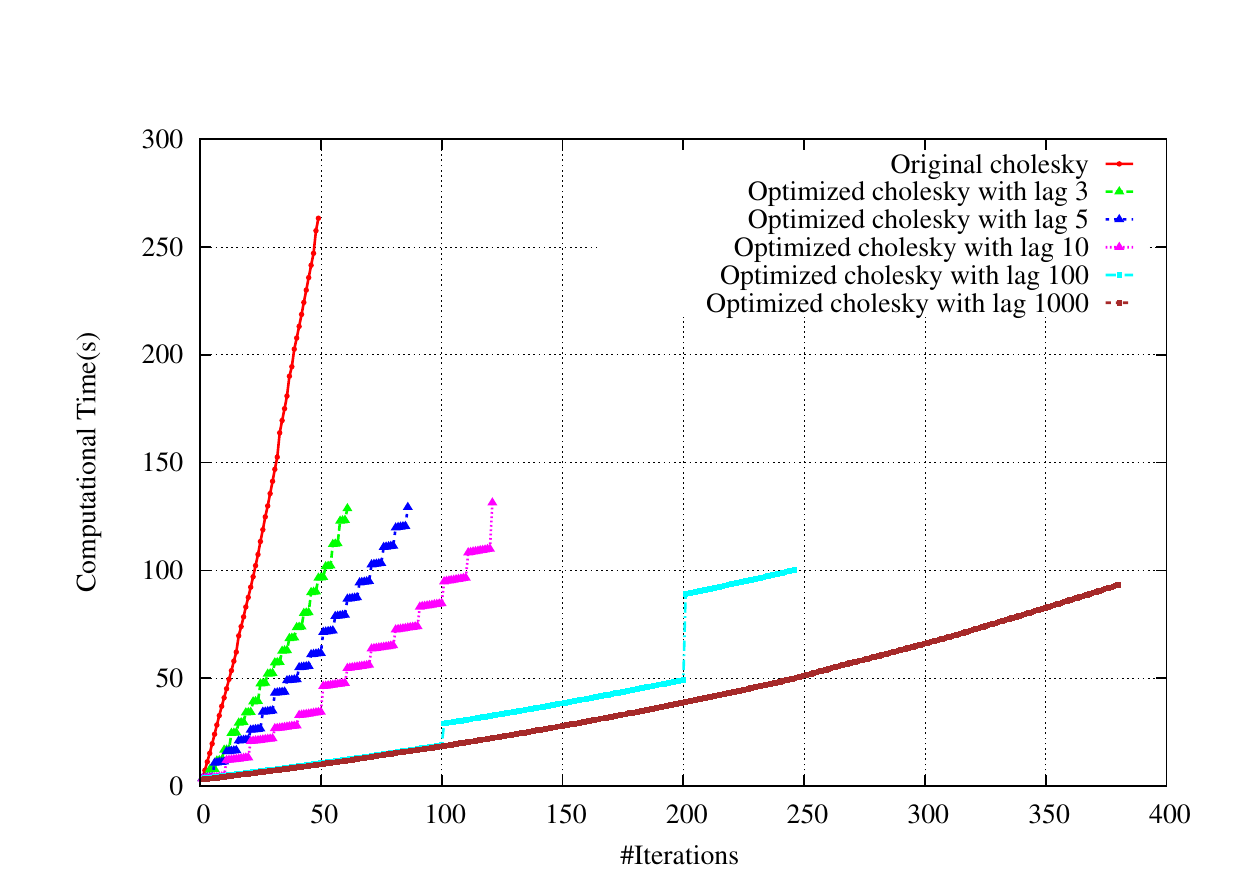}
  \caption{Comparison of computational time and number of iterations until a fixed accuracy of original Cholesky factorization and our approach (Optimized Cholesky factorization) using $5$ dimensional levy function for different lags using $200$ seeds}
  \label{fig:chol_seed_200_difflag}
\end{figure*}

The $d$-dimensional Levy function is a well-suited test case to evaluate the performance of an optimization approach \cite{levy_function} and defined as 
\begin{align}
\begin{split}
 f(\textbf{x}) = &\sin^2(\pi w_1) \\
    &+ \sum_{i=1}^{d-1} (w_i-1)^2 [ 1+10 sin^2(\pi w_i+1)]  \\ 
    &+ (w_d-1)^2  [1+sin^2 (2\pi w_d)]  \\ 
    &\text{where} \ w_i = 1 + \frac{x_i- 1 }{4}, \  \text{for}  \ i = 1, \ldots, d
\end{split}
\end{align}
In our setting, we use $d=5$ to generate a function reasonable difficult to maximize; see Fig. \ref{fig:levy} to get an impression of this test function. Please note, that we decided to maximize the negative Levy function to be in accordance to our remaining experiments. 
Typically, the negative Levy function is evaluated on the hyper-cube  $x_i \in [-10, 10]$, $\forall i \in [1,d]$ while its global maximum is ($f(\textbf{x}^{*}) = 0$) at  $x^{*} = {(1, \ldots ,1)}$.\\
We investigate the behavior of our approach in comparison to the original one with two main aspects in mind: First, we analyze the computation time per iteration. The Cholesky decomposition used to be the bottleneck of Bayesian optimization with Gaussian processes. We evaluate therefore the impact of our adjustment to overall run-time, see Fig. \ref{fig:chol_vs_iter}.  We computed 100 iterations for both approaches and plot the corresponding computation times in the log-scale. We observe an asymptotic behavior of $\mathcal{O}({n^3})$ and $\mathcal{O}({n^2})$ for the original and the optimized Cholesky factorization, respectively. With the increasing number of iterations, the high computational speed of our approach outperforms the original one with a total factor of about $162$. \\
The second important aspect of every hyperparameter optimization method is the time to convergence. Thus, we make 1000 iterations with both methods and evaluate whether and if so when the optimal solution, i.e. $f(\bm{x})=0$ is approximated reasonable well; see Tab. \ref{tab:maxiter_levy}. We evaluated two settings: In the first one, we provided only 1 random seed and continued for 1000 iterations of the optimization procedure. In the second one, we generated 100 random seeds and used this as a broad initialization for the remaining 1000 iterations.
\begin{table}[t]
    \centering
    \begin{tabular}{c|c|c|c}
    \hline
  \noalign{\smallskip}
  \multicolumn{4}{c}{\textbf{Na\"ive Cholesky decomposition}}\\
  \noalign{\smallskip}
  \hline
  \noalign{\smallskip}
  \multicolumn{2}{c}{1 seed point}
  & \multicolumn{2}{c}{100 seed points}\\
  \noalign{\smallskip}
  \hline
  \noalign{\smallskip}
        Iteration & Accuracy & &\\
        \noalign{\smallskip}
\hline
\noalign{\smallskip}
        82 & -5.23 & 168 & -0.52 \\
        126 & -4.5 & 169 & -0.41\\
        386 & -4.3 & 170 & -0.30\\
        479 & -4.1 & \textbf{232} & \textbf{-0.04} \\
\noalign{\smallskip}
\hline
  \noalign{\smallskip}
  \multicolumn{4}{c}{\textbf{Optimized Cholesky decomposition}}\\
  \noalign{\smallskip}
  \hline
  \noalign{\smallskip}
  \multicolumn{2}{c}{1 seed point}
  & \multicolumn{2}{c}{100 seed points}\\
  \noalign{\smallskip}
  \hline
  \noalign{\smallskip}
  Iteration & Accuracy & &\\
  \noalign{\smallskip}
\hline
\noalign{\smallskip}
        103 & -1.3 &  648 & -0.95\\
        139 & -0.77 & 673 & -0.61\\
        156 & -0.33 & 731 & -0.23\\
        297 & -0.18 & 838 & -0.19\\
        318 & -0.14 & 972 & -0.18\\
        \textbf{611} & \textbf{-0.01} & 975 & -0.09\\
\noalign{\smallskip}
\hline
\noalign{\smallskip}
    \end{tabular}
    \caption{Accuracy improvements of our approach and the na\"ive Cholesky decomposition for the 5D Levy function.}
    \label{tab:maxiter_levy}
\end{table}
Please note that both test settings are fundamentally different. In the first one, almost nothing is known about the function and the complete optimization is based on the suggested samples. In the latter case, however, a rough approximation of the target function is already contained in the basic data by random sampling - only the optimum position still has to be hit.\\
If there are almost no seeds available, we observe the well-known problem for the standard version of the Bayes optimization with expected improvement: The acquisition function gets trapped in a local maximum and is not able to identify the ideal parameter combination. In contrast, our approach converges to the global maximum in 611 iterations.\\
In the scenario of 100 seed points, our approach still converges to the optimal parameter setting, but needs more iterations - we do not learn the shape of the covariance kernel from the data when we do not update the parameters of the covariance matrix. Unfortunately, this fixation is necessary to re-use the precomputed Cholesky factor $\mathbf{L}$. \\
To combine our approach with the full factorization, we generate a new setting and update the covariance matrix after a fixed number of iterations, which we call the lagging factor $l$. If this factor equals one, we resemble the full factorization every iteration, i.e. our approach equals the standard variant.
Fig. \ref{fig:chol_seed_200_difflag} shows the effect of lagging the kernel updates on computational time and the number of iterations until convergence for a fixed accuracy. With an increasing lagging factor, the number of iterations to convergence also increases while the computational time decreases dramatically. Please note, that the jumps in the computation time are related to the full matrix inversions necessary after an update step of the covariance kernel function. Similarly, the less we update the kernel parameters, the faster is the computation, i.e. the complexity of our approach tends to $\mathcal{O}({n^2})$ for $l \rightarrow \infty$. \\
We empirically investigated the lagging factor and set $l=3$ in our experiment. With the same number of seeds, our approach reaches a close to the optimal solution within 192 iterations and a function evaluation of $-0.21$ to the optimum.
All in all, these experiments indicate, that our approach does not get trapped in local maxima in contrast to the original method. However, as soon as a reasonable number of seeds providing a coarse approximation to the underlying function is given, the kernel parameters of the covariance function should be updated at reasonable intervals. 
Nevertheless, in the scenario of hyperparameter optimization of neural networks, such seed points are very costly: training runs can take even days. In the following, we investigate our approach for two different network architectures on two different data sets.
\subsection{LeNet and MNIST}
The MNIST dataset is one of the classic benchmarks for classification models \cite{lecun1998mnist} and contains ten classes of handwritten digits with corresponding labels. Although it has been possible to solve this benchmark very reliably and accurately since the release of LeNet \cite{lecun1990handwritten}, it still has some advantages: First, it is an understandable and intuitive task. Second, the data set is relatively small, which allows for fast training of deep neural networks in average $8$ seconds for 10 epochs.\\
MNIST data set contains 60000 images of size $28\times 28$. In our experiments, we use 48000 images for training, 6000 for test, and 6000 for validation, respectively. We train the LeNet5 network including 2 dropout layers after the fully connected layers with stochastic gradient descent and momentum and a batch size of 128 \cite{lecun1995comparison,robbins1951stochastic,qian1999momentum, srivastava2014dropout}. The hyperparameters we optimized are the keep probability for the dropout layers $d_1,d_2 \in [0.01,1] $, learning rate $lr \in [0.0001, 0.1]$, weight decay $w\in [0,0.001]$, and momentum $m \in [0,0.99]$.\\
We compare the original implementation (see Alg. \ref{algo:Cholesky}) with our approach (without any updates of the kernel parameter) in two main aspects: First, we analyze the computational overhead for network training in Fig. \ref{fig:mnist_comp_overhead}. \\
While our approach does generate only minor additional computation time, the overhead of the exact Cholesky factorization results in massive time consumption. In the beginning, both approaches perform similarly in terms of speed. However, as the matrix grows and the corresponding decomposition time explodes cubically, the overhead dominates the hyperparameter optimization procedure. After 1000 iterations, the na\"ive approach still needs the same time for training, but in total 4.5 times as long per iteration in comparison to the first steps in the optimization. In contrast, our approach stays stable concerning time per iteration. Although the matrix increases, the computational burden is still feasible resulting in no noticeable overhead.\\
\begin{table}[b!]
    \centering
    \begin{tabular}{c|c}
    \hline
  \noalign{\smallskip}
  \multicolumn{2}{c}{\textbf{Na\"ive Cholesky decomposition}}\\
  \noalign{\smallskip}
  \hline
  \noalign{\smallskip}
        Iteration & Accuracy \\
        \noalign{\smallskip}
\hline
\noalign{\smallskip}
        13 & 0.25 \\
        15 & 0.67 \\
        24 & 0.83 \\
        26 & 0.88 \\
        63 & 0.90 \\
        198 & 0.93 \\
        239 & 0.96 \\
        732 & \textbf{0.97} \\
\noalign{\smallskip}
\hline
  \noalign{\smallskip}
  \multicolumn{2}{c}{\textbf{Optimized Cholesky decomposition}}\\
  \noalign{\smallskip}
  \hline
  \noalign{\smallskip}
  Iteration & Accuracy \\
  \noalign{\smallskip}
\hline
\noalign{\smallskip}
        1 & 0.11 \\
        4 & 0.12 \\
        11 & 0.96 \\
        \textbf{168} & \textbf{0.97} \\
\noalign{\smallskip}
\hline
\noalign{\smallskip}
    \end{tabular}
    \caption{Accuracy improvements of our approach and the na\"ive Cholesky decomposition for LeNet5 and MNIST.}
    \label{tab:maxiter_mnist}
\end{table}
However, speedup is not important when the optimization procedure does not find reasonable solutions. Hence, we also compared both approaches in terms of test accuracy; see Tab. \ref{tab:maxiter_mnist}. 
Our approach does not suffer from the lower performance. Indeed, our method can find the optimal solution in 168 iterations, while the na\"ive approach needs 732 optimization steps. This results in an improved computation time of $24.6$min in comparison to $372.47$min for the original approach and therefore in a dramatical performance boost of $\approx 93 \%$ and a speedup of factor 15. The comparable performance indicates that our relaxation does not harm the optimization strategy when it comes to the training of deep neural networks. 
To investigate this indication even further, we trained ResNet32 on CIFAR10, a widely used benchmark data set.

\subsection{ResNet and CIFAR10}
CIFAR10 \cite{krizhevsky2009learning} is a frequently used benchmark data set, composed of 60000 images of size $32 \times 32$ from ten classes. Here, 50000 images with 5000 samples of each class are in the training, and 10000 in the test set. It is a smaller version of CIFAR100, with 100 different classes. Typically, hyperparameters are optimized on the smaller data set and afterward applied to the larger one.\\
We train ResNet32, a standard deep learning architecture, for this classification problem. Similar to before, we use stochastic gradient descent with momentum and weight decay and set the batch size to 128 for 10 epochs. The hyperparameters we optimized are learning rate $lr \in [0.0001, 0.1]$, weight decay $w\in [0,0.001]$, and momentum $m \in [0,0.99]$.\\
In contrast to LeNet5, network training is more time-intense with $190$ sec on average. A high sample efficiency is therefore even more important. However, the bottleneck of the Cholesky factorization does not occur in this setting as training time dominates the overall duration per iteration. Similar to our previous results, our approach identifies a better parameter setting resulting in an accuracy of 0.81 after 10 epochs, see Tab. \ref{tab:maxiter_cifar}. We reach a competing parameter setting to the na\"ive approach on average already after 62 iterations in $194.2$min - the original algorithm takes on average 176 iterations with a duration of $567$min, i.e. ours is in total 3 times faster.  Nevertheless, the ability for strong scalability is essential in this scenario. 
The results indicate, that our approach is more robust against local maxima while being much more sample efficient.
\begin{table}[t!]
    \centering
    \begin{tabular}{c|c}
    \hline
  \noalign{\smallskip}
  \multicolumn{2}{c}{\textbf{Na\"ive Cholesky decomposition}}\\
  \noalign{\smallskip}
  \hline
  \noalign{\smallskip}
        Iteration & Accuracy \\
        \noalign{\smallskip}
\hline
\noalign{\smallskip}
        16 & 0.74 \\
        26 & 0.75 \\
        43 & 0.77 \\
        50 & 0.78 \\
        176 & 0.79 \\
\noalign{\smallskip}
\hline
  \noalign{\smallskip}
  \multicolumn{2}{c}{\textbf{Optimized Cholesky decomposition}}\\
  \noalign{\smallskip}
  \hline
  \noalign{\smallskip}
  Iteration & Accuracy \\
  \noalign{\smallskip}
\hline
\noalign{\smallskip}
        24 & 0.77 \\
        53 & 0.78 \\
        62 & 0.79 \\
        117 & 0.80 \\
        237 & \textbf{0.81} \\
\noalign{\smallskip}
\hline
\noalign{\smallskip}
    \end{tabular}
    \caption{Accuracy improvements of our approach and the na\"ive Cholesky decomposition for ResNet32 and CIFAR10.}
    \label{tab:maxiter_cifar}
\end{table}
To investigate these findings even further, we also experimented with the parallelization of the network training.
\subsection{Parallelized ResNet and CIFAR10}
The training procedure of deep neural networks but even more the topology search of neural networks' architectures critically depend on a proper tuning of hyperparameters. As a first step towards the parallelization of hyperparameter optimization, we set up an experiment to see whether our approach scales properly in a high-performance context.
Thus we follow our strategy of Sec. \ref{sec:parallelization} and do not only use the maximal expected improvement overall seeds but the 20 best local maxima; see Sec. \ref{sec:ei}.\\
First, our approach scales well and we improve dramatically in overall run-time in comparison to the original approach: While the original approach needs 176 iterations, we hit the same accuracy after 35 optimization steps, i.e. a speed up with a factor of 5.
We even outperform our approach and reach our final result of 0.80 after 61 iterations, i.e. $\approx 50\%$ less time than our sequential method.

\section{Conclusions}
\label{sec:conclusions}
We have investigated the use of lazy Gaussian processes, which allow enormous accelerations in the field of hyperparameter optimization for artificial neural networks. As a first contribution, we have shown that neglecting the iterative parameter updates of the covariance kernel function can prevent Gaussian processes with EI from being trapped in local extrema. \\
We then used this relaxation to solve the Cholesky factorizations iteratively using the last calculated factor. This has led to the fact that we could prove from simple problems like the optimization of the Levy function, overtraining of a relatively simple network, up to training of modern architectures, an enormous reduction of the running time. In the first scenario, we demonstrated a speed up with factor 162, while hyperparameter optimizations of neural networks showed speed ups up from 5 to 15 for ResNet32 and LeNet5, respectively.\\
Our target functions for hyperparameter optimization of deep neural networks are still too simple: The hyperparameter space is not large enough, so we do not need enough iterations to show the full power of our approach. However, our method is especially well suited for massive parallelization of network training; see Fig. \ref{fig:mnist_comp_overhead}: We could shrink the bottleneck of synchronization - updating the Gaussian posterior - to a minimal computational overload. In our current research, we are focusing on neural architecture search where the domain of possible hyperparameters increases dramatically, and therefore the need for high numbers of samples explodes.

\begin{table}[t!]
    \centering
    \begin{tabular}{c|c}
    \hline
  \noalign{\smallskip}
  \multicolumn{2}{c}{\textbf{Optimized Cholesky decomposition}}\\
  \noalign{\smallskip}
  \hline
  \noalign{\smallskip}
  Iteration & Accuracy \\
  \noalign{\smallskip}
\hline
\noalign{\smallskip}
        1 & 0.11 \\
        16 & 0.51 \\
        21 & 0.77 \\
        35 & 0.79 \\
        61 & \textbf{0.80} \\
\noalign{\smallskip}
\hline
\noalign{\smallskip}
    \end{tabular}
    \caption{Accuracy improvements of our approach decomposition for ResNet32 and CIFAR10 in a parallel setting.}
    \label{tab:maxiter_cifar_par}
\end{table}
\section*{Acknowledgment} 
The authors gratefully acknowledge funding of the DeToL project by the Federal Ministry of Education and Research of Germany as well as Dr. Daniel Gr\"unewald for fruitful discussions.

\bibliographystyle{unsrt}
\bibliography{./bibliography/arxiv} 
\end{document}